\RequirePackage{fix-cm}
\documentclass[smallextended]{svjour3}       
\smartqed  
\usepackage{graphicx}
\usepackage{booktabs}  
\usepackage{amsmath}
\usepackage{textcomp}  
\usepackage{lipsum}    

\usepackage[table]{xcolor}
 
\newcommand\MyBox[2]{
  \fbox{\lower0.75cm
    \vbox to 1.7cm{\vfil
      \hbox to 1.7cm{\hfil\parbox{1.4cm}{#1\\#2}\hfil}
      \vfil}%
  }%
}
\definecolor{burgundy}{rgb}{0.5, 0.0, 0.13}
\usepackage{enumitem}

\usepackage{longtable}
\usepackage{float}
\usepackage{multirow}
\usepackage{booktabs}

\usepackage{subcaption}
\usepackage{cite}
\usepackage[tableposition=top]{caption}
\usepackage{diagbox}

\usepackage{graphicx}

\usepackage{changepage}
\begin{document}

\title{Enhancing a Student Productivity Model for Adaptive Problem-Solving Assistance
}


\author{
{\large Mehak Maniktala} 
\and  
{\large Min Chi}
\and 
{\large Tiffany Barnes}
}


\institute{Mehak Maniktala \at
              \email{mmanikt@ncsu.edu}           
           \and
           Tiffany Barnes \at
              \email{tmbarnes@ncsu.edu}
}

\date{Received: date / Accepted: date}

\maketitle

\begin{abstract}
Research on intelligent tutoring systems has been exploring data-driven methods to deliver effective adaptive assistance. While much work has been done to provide adaptive assistance when students seek help, they may not seek help optimally. This had led to the growing interest in proactive adaptive assistance, where the tutor provides unsolicited assistance upon predictions of struggle or unproductivity. Determining \emph{when} and \emph{whether} to provide personalized support is a well-known challenge called the assistance dilemma. Addressing this dilemma is particularly challenging in open-ended domains, where there can be several ways to solve problems. Researchers have explored methods to determine when to  proactively help students, but few of these methods have taken prior hint usage into account. In this paper, we present a novel data-driven approach to incorporate students’ hint usage in predicting their need for help. We explore its impact in an intelligent tutor that deals with the open-ended and well-structured domain of logic proofs. We present a controlled study to investigate the impact of an adaptive hint policy based on predictions of HelpNeed that incorporate students’ hint usage. We show empirical evidence to support that such a policy can save students a significant amount of time in training, and lead to improved posttest results, when compared to a control without proactive interventions. We also show that incorporating students’ hint usage significantly improves the adaptive hint policy’s efficacy in predicting students’ HelpNeed, thereby reducing training unproductivity, reducing possible help avoidance, and increasing possible help appropriateness (a higher chance of receiving help when it was likely to be needed). We conclude with suggestions on the domains that can benefit from this approach as well as the requirements for adoption.

\end{abstract}

\section{Introduction}
Personalized instruction is considered an effective educational strategy \cite{bloom19842}, with research suggesting that students who receive this individualized instruction perform significantly better than those who receive classroom instruction \cite{ma2014intelligent, vanlehn2011relative}. However, human tutors may not always be available or accessible to each student at all times which, in turn, makes personalized human tutoring infeasible at a large scale. This has led to the growing success of intelligent tutoring systems (ITSs) because they can provide \textit{user-adaptive} instruction and feedback at a large scale \cite{kardan2015providing, ueno2017irt}.

Research suggests that hints when used appropriately can augment students’ learning experience and improve their performance \cite{bunt2004scaffolding,bartholome2006matters}, but they may not seek help optimally \cite{aleven2006toward}. One way to tackle non-optimal help-seeking behaviors, such as help avoidance, is to provide unsolicited help. The \textit{assistance dilemma} is a trade-off between information giving and withholding to achieve optimal learning \cite{koedinger2007exploring}. A core problem of the assistance dilemma is the need to discover \emph{when} and \emph{whether} students are unproductive so that the tutor can intervene. Modeling student behavior for adaptive unsolicited assistance in ITSs, especially that deal with open-ended domains, is a well-recognized challenge \cite{mclaren2014web, borek2009much}. This is because open-ended domains have problems that can be solved in several ways, making it difficult to figure out what path students are on, and how to adapt assistance for effective learning. 

Open-ended domains can be ill-structured, where problems do not have a clear goal, set of operations, end states, or constraints; or they can be well-structured, where problems have a clear goal, end states, or constraints. The assistance dilemma has been challenging to address in open-ended domains that deal with both ill-structured problems \cite{mclaren2014web,borek2009much} and well-structured problems \cite{ueno2017irt}. In our prior work, we developed a data-driven approach to solve the assistance dilemma in Deep Thought, an intelligent tutor for the open-ended and well-structured domain of logic proofs. While logic problems are well-structured in that they contain all the information needed to solve the problem and there are well-defined algorithms that students can use to solve them, they are open-ended in the sense that they have many possible solutions that can all be correct. We developed a model of student productivity, and used it to form an \textit{adaptive hint policy} that proactively intervenes upon predictions of \textit{HelpNeed} or unproductivity \cite{maniktala2020jedm}. While the controlled study showed promising results, it also showed that our HelpNeed predictor had a higher than expected proportion of false negatives (student steps that were falsely predicted to not need help, when actually they did). We hypothesized that the predictor was highly biased towards efficient steps and was unable to differentiate efficient steps carried out with tutor help, from those without. So, in this paper, we seek to examine and understand the impact of incorporating student hint usage in modeling their productivity, and consequently, the impact of the \textit{revised} adaptive hint policy on student learning and performance. 

We present a novel method to incorporate student hint usage in modeling their HelpNeed, and an analysis to investigate its impact on HelpNeed predictions. We also present a controlled study that showcases the significance of providing Adaptive assistance proactively upon predictions of HelpNeed. Our results show that, unlike our prior adaptive hint policy (that did not take into account students’ hint usage), students in this study’s Adaptive condition have significantly lower unproductive HelpNeed steps in training than the Control. We also observe that the Adaptive condition has lower possible help avoidance, and higher possible help appropriateness (a higher chance of receiving help when it was likely to be needed) than the Control. The improved training productivity of the Adaptive condition saves them about half an hour on average (about 20\% speedup) and enables them to perform better on a posttest when compared to the Control.

The main contribution of this paper is the simple yet effective method of incorporating students’ hint usage in modeling their training productivity. Further, while much research has explored ways to incorporate help usage to improve student models, to the best of our knowledge, no prior work has investigated and contrasted the impact of delivering proactive assistance in a well-structured open-ended domain using student modeling with and without incorporating students’ help usage. Our second contribution is an extensive analysis of a controlled study carried out to investigate the impact of providing adaptive hints proactively using a HelpNeed predictor that takes into account students’ help usage, and contrasting it with a predictor that does not.

In the remainder of the paper, we survey the related work. Next, we elaborate the HelpNeed predictor and how we incorporate students’ hint usage. We present an Adaptive hint policy that uses the revised HelpNeed predictor in a controlled study, its impact on students’ training productivity and posttest performance. We also investigate whether incorporating students’ help usage helps solve the assistance dilemma more effectively than our prior HelpNeed study.

\section{Related Work}
\subsection{Data-driven User-Adaptive Assistance}
Researchers have extensively explored data-driven methods to provide adaptive support to students in ITSs. Data-driven assistance has shown to save time and resources by reducing the need for an expert to construct hints for every possible student situation \cite{murray2003overview}. For example, consider the task of generating hints in the domain of logic proofs. Deep Thought has N = 72,560 unique problem-solving states in the prior student data. It would be infeasible for an expert to generate a hint for each of these states. Instead, we use the Hint Factory, a data-driven approach to generate next-step hints \cite{barnes2008pilot}, and our recent modification to it guarantees a 100\% hint availability \cite{maniktala2020leveraging}.

While data-driven hint generation is crucial in providing adaptive assistance for open-ended domains, an equally important tutoring aspect is to determine \textit{when} to administer help. Much research on hints has explored the impact of \textit{on-demand} demand hints, i.e., when students request them \cite{razzaq2010hints}. Since students may not seek hints optimally, several studies have explored the impact of \textit{agency}, i.e., whether hints are requested by the student or proactively provided by the tutor. While some studies suggest that student-initiated on-demand hints lead to better learning \cite{razzaq2010hints}, other studies suggest that tutor-initiated proactive help is better \cite{arroyo2001analyzing, murray2006comparison}. The conflicting results from different studies highlight the need to explore tutor-initiated assistance in more detail, especially because educational psychology suggests that tutor is an active participant in scaffolding \cite{wood1976role}, where learning first occurs at a social level \cite{vygotsky1980mind, stone1993missing} when a tutor brings a concept within students’ zone of proximal development \cite{vygotsky1980mind}. A study by Brawner on unsolicited assistance showed that while humans naturally intervene when students need help, it is not as easy to incorporate in ITSs, and that unsolicited assistance is relatively unexplored for ITs because they need additional resources \cite{brawner2011understanding}. The following section details the challenge in determining \textit{when} ITSs should intervene, and provides examples of how researchers have employed student modeling to do so.

\subsection{Assistance Dilemma}
According to Koedinger et al., assistance dilemma is about determining the amount and timing of help to achieve optimal student outcomes \cite{koedinger2007exploring}. While withholding more information than needed can lead to frustration and wasted time, giving more information than needed can lead to shallow learning and a lack of motivation to learn by oneself \cite{koedinger2007exploring, mclaren2014web}. Koedinger and Aleven \cite{koedinger2007exploring} worked towards addressing this dilemma in a cognitive tutor by initially withholding information about problem solutions and steps, and then interactively adding information, only as needed, through yes/no feedback, explanatory hints, and dynamic problem selection. McLaren et al. explored solving the assistance dilemma in an ill-structured inquiry-based chemistry tutor \cite{mclaren2008and}. They experimented with three levels of problem-level assistance: high (worked examples - tutor showing sample solutions), mid-level (less assistance than worked examples), and low (untutored problem solving). Their study suggests that, among the three levels, mid-level assistance leads to the highest learning. 
 
Researchers have also used student modeling techniques to address the assistance dilemma  \cite{murray2004looking,ueno2017irt,conati2002using,kock2010towards}. For example, Conati et al. used Bayesian Networks to model student behavior in Andes, a physics tutor, to, in part, determine when to provide students with unsolicited mini-lessons \cite{conati2002using}. In another study, Kardan and Conati modeled student behavior using clustering and association rule mining for determining when to provide unsolicited adaptive hints in an exploratory interactive tutor for constraint satisfaction problems \cite{kardan2015providing}. They found that the adaptive support helped students to have significantly higher learning than those without, but it did not lead to improved task performance. Fratamico et al. applied Kardan and Conati’s 2015 framework in an electronic circuits simulator and found that it successfully classified students into groups of high and low learners \cite{fratamico2017applying}. In our prior work on solving the assistance dilemma in a logic tutor, we introduced a data-driven method to determine and predict unproductive or HelpNeed steps. We used this HelpNeed predictor to determine when to intervene to provide partially worked steps. Our findings suggest that such interventions enabled students to form shorter (more efficient) solutions in a posttest in less time than those without proactive interventions \cite{maniktala2020jedm}. 

\subsection{Modeling Unproductivity}
Unproductivity refers to undesirable training behavior that is often associated with poor performance \cite{kai2018decision,botelho2019developing}. One of the most widely used definitions of unproductive behavior was introduced by Beck and Gong \cite{beck2013wheel}. They define wheel-spinning or unproductive persistence based on whether a student achieves mastery (three correct problems) in ten problem attempts. This definition has been used in recent studies to predict unproductive behavior with decision trees \cite{kai2018decision} and recurrent neural networks \cite{botelho2019developing}. Despite its widespread use, this definition is not appropriate for our task of determining step-level productivity (to provide step-level assistance) because it assesses students on a problem-level. Even on a problem-level, this definition is not suitable for open-ended problems that value shorter, more efficient solutions.

McLaren in a study on an ill-structured open-ended chemistry tutor defined unproductive events as actions that are unlikely to advance students’ understanding \cite{mclaren2014web}. In our prior work, we developed a model of productivity to identify problem-solving steps that are not likely to advance students’ problem-solving strategies \cite{maniktala2020extending}. But our definition is different from McLaren’s definition because we do not use a predefined domain-specific metric for efficient and inefficient strategies. 

Efficient strategies in many open-ended multi-step domains (both ill- and well-structured) are reflected in shorter solutions with less problem-solving time \cite{mayer1992thinking,smith2012toward}. Our model of productivity extends the data-driven Hint Factory approach \cite{maniktala2020jedm} to determine efficient steps, and uses pedagogical theory to combine step efficiency and duration for determining HelpNeed, i.e, steps where students need help. More details about our HelpNeed approach relevant to this study are presented in section \ref{sec:f19_policy} below.

\subsection{Hint Usage in Student Modeling}
This work assesses the impact of incorporating hint usage in predicting student behavior. While researchers have extensively explored predictive student modeling tasks such as predicting student performance \cite{wang2011assistance, chaudhry2018modeling}, problem selection \cite{mostafavi2015data}, and adaptive assistance \cite{conati2002using}, less attention has been paid to the role of students’ hint usage in such modeling tasks. The work by Beck et al. in \cite{beck2004automatically} was one of the first to assess the impact of tutor help on predictive modeling. They showed that taking into account students' help requests improved the predictive power of a reading tutor. Later, they developed the Help model \cite{beck2008does}, an extension of a basic knowledge tracing model \cite{corbett1994knowledge}, to explicitly model the role of help in student knowledge and performance. This work was the first to jointly model tutor intervention and student knowledge, and learn the parameters of a Bayesian network by incorporating students’ help usage. Most work on knowledge tracing models before this either ignored students' hint intake or labelled hint intake as incorrect response \cite{beck2008does}. A recent study by Chaudhry et al. developed a multi-task memory augmented deep learning model to jointly predict students’ hint-taking and knowledge tracing tasks \cite{chaudhry2018modeling}. Their proposed model significantly outperformed other models that do not take into account hint usage by at least 12\%. 

One group has conducted several studies evaluating the role of help-seeking in predicting student performance in ASSISTment, a math tutor \cite{feng2009addressing, wang2010representing, wang2011assistance, hawkins2013extending}. One of their earlier works by Feng et al. dealt with building regression models using student-tutor interaction data to estimate student performance on a high-stakes state test \cite{feng2006predicting}. They found a significant and negative correlation between hint requests and test score. Later, they found more evidence to suggest that the online assessment system can do a better job of predicting student knowledge by taking into account the tutoring assistance needed \cite{feng2009addressing}. Another study by Wang and Heffernan showed evidence to support that incorporating students’ hint requests and attempts contributes to more predictive power than binary performance, and suggested that incorporating such features can have the potential to enhance student modeling techniques \cite{wang2010representing}. They also introduced the Assistance Model (AM) for predicting student performance using information about the number of hints and attempts a student needed to answer the previous question \cite{wang2011assistance}. They showed that, while this model alone did not lead to better prediction performance than other models, ensembling it with other models reliably improves the predictive accuracy. 

A study by Emerson et al. on predictive student modeling for PRIME, a block-based programming environment, mentions that there is a need to develop predictive models that can accurately identify negative student behaviors to evolve beyond providing on-demand hints \cite{emerson2019predicting}.They created a predictive model for student-activity completion informed by four families of features including hint usage, prior performance, activity progress, and interface interaction, and found that it consistently outperformed their baselines. 

While the above mentioned studies evaluated the role of hint usage in either predicting student performance or attrition, they do not show the effect of using such models to proactively intervene. In this study, we present both a method to incorporate hint usage for modeling student productivity in a well-structured open-ended domain, and a controlled study that investigates the impact of proactive interventions using this model.

\section{HelpNeed Policy}
\label{sec:f19_policy}
In this section, we briefly discuss the HelpNeed policy -- why it was built, how it was built, its impact on students, and drawbacks.

The HelpNeed policy was developed for Deep Thought, a data-driven intelligent logic tutor where students practice constructing multi-step propositional logic proofs in discrete math courses, with 200-350 students per semester, since 2008. Our prior work shows that even though Deep Thought provides hints with user-adaptive content, students, in general, avoided seeking help \cite{maniktala2020leveraging}. To reduce help avoidance and address the assistance dilemma in Deep Thought, we built an adaptive hint policy \cite{maniktala2020jedm}, which at the start of each problem-solving step uses a HelpNeed (HN) predictor to predict if students will need help learning efficient problem-solving strategies in the next step. If the predictor predicts HelpNeed, the adaptive hint policy intervenes with a proactive hint. This proactive hint is a partially worked step that provides students information on the next, most optimal, logic statement they can derive to move towards the solution. To use the proactive hints, students must \textit{justify} or derive the hinted statement by selecting the appropriate existing statements and logic rule. 

The HelpNeed definition is based on step efficiency and duration. \textit{Efficiency} is our data-driven metric that reflects a step’s quality and how well it promotes progress to a good solution. We provide more details on step efficiency relevant to this study in section \ref{sec:s3_step_eff}. \textit{Duration} is the time taken by a student to carry out a step, which in the HelpNeed model is said to be long if it is above the threshold of the $75^{th}$ percentile, or quick otherwise.

In our prior HelpNeed policy \cite{maniktala2020jedm}, we defined five step behaviors (from most expert to least expert): (1) \textit{Expert-like}: quick efficient step, (2) \textit{Strategic}: long efficient step, (3) \textit{Opportunistic}: one quick inefficient step, (4) \textit{Far Off}: consecutive quick inefficient steps, and (5) \textit{Futile}: long inefficient step. Among these five, \textit{Far Off and Futile are HelpNeed steps} as prior research suggests that students can benefit from help either when they are far off track \cite{borek2009much} or when they spend a long time without making progress towards learning the concepts being taught \cite{beck2013wheel, kai2018decision}. Note, \textit{we do not consider Opportunistic steps as needing help} because they represent guess-and-check opportunistic strategy and literature suggests that students should be given opportunities to use guess-and-check strategies and they can learn from effectively checking their guesses in semi open-ended domains \cite{polya2004solve} but prolonged guessing without progress needs intervention \cite{kinnebrew2014analyzing}.

Next, we briefly describe our HelpNeed predictor. This predictor  has two random forest binary classifiers: \textit{state-based} and \textit{state-free}, with class 1 representing HelpNeed. A problem-solving \textit{state} is a snapshot of students’ on-going or complete proof at any given time. Each logic statement derivation or deletion causes a change in the state, and this transition between states is called a step. We use the \textit{state-based} classifier when a student's problem-solving state can be matched to historical data (logs recorded for 796 students) to leverage more informative data-driven step efficiency related features for predictions (see section \ref{sec:s3_step_eff}). However, since the domain of logic is open-ended, we may not always find a state-match. The \textit{state-free} classifier is used when we don’t have that information. This two-classifier architecture, described in more detail in \cite{maniktala2020jedm}, ensures that a HelpNeed prediction can be made regardless of whether or not we find a state match.

We found that the HN predictor and the adaptive hint policy based on it shows promising results in a controlled study \cite{maniktala2020jedm}, where students in the HN Adaptive condition, who receive adaptive hints based on the HN model, perform better in the posttest, and form significantly more optimal proofs in less time, than those in the HN Control condition who do not receive this proactive intervention. The HN Adaptive condition shows significantly less help avoidance and abuse than the control. However, the HN predictor has a higher proportion of false negatives for students in the HN Adaptive condition than the Control. We hypothesize that the predictor is strongly biased towards efficient steps, and when students incorporate the proactive hints in their proofs, i.e. they \textit{use} the hints, their solutions appear more expert-like, which fools the predictor into inferring that more interventions are not needed, even when they are. In this study, we incorporate students’ hint usage in the HN predictor to create a HelpNeed and Use (HNU predictor). To do so, we revised how we compute step efficiency to account for hint usage in our predictive modeling. We first detail the step efficiency metric in the next section and then describe how we incorporate hint usage to predict HelpNeed.

\subsection{Step Efficiency} 
\label{sec:s3_step_eff}
This section provides a brief review of the data-driven step efficiency metric we defined in our prior HelpNeed study \cite{maniktala2020jedm}. Step efficiency is an extension of the Hint Factory \cite{barnes2011using} that measures whether a student’s most recent step contributes to an efficient (short) solution. 

Since the domain of logic is open-ended, students can follow a variety of solution paths at any given problem-solving state. While some paths may not lead to solutions, some may lead to a solution that can be highly inefficient. Further, while a student can be in a state that leads to an efficient solution, there may be low probability for the student to select that path. Our step efficiency method takes these aspects into account and has two parts. First, an interaction network is generated using prior student data, where each node is a state, each edge is a step (state transition), and the probability of each state-transition is recorded. A state transition occurs upon deriving a new logic statement or deleting one from the proof. Next, we carry out the Bellman backup for value iteration (used in reinforcement learning) on the states to determine their \textit{quality} values. This involves assigning large rewards to solution states, large penalties to states that never lead to solutions, and small penalty for carrying a step (to penalize longer solutions). Note that the Bellmann backup also considers the probability of transition between states while assigning state quality values. Our prior work defined two types of state quality metrics - \textit{local} which compares states one step away, and \textit{global} quality compares all states \cite{maniktala2020jedm}. 

As described in our prior work, each problem can have a different range of state quality values (dependent on its state-space and difficulty). So, to measure step efficiency, we need a reference state against which we can compare a students’ current state quality. There can be two reference states - the previous state, in which case we say that we are measuring \textit{relative progress}, or the start state, in which case we measure the \textit{absolute progress}. A step is called efficient if the progress, using either quality metric, is a non-negative number. Since there are two ways to measure quality and two ways to measure progress, we investigated the use of all four combinations to define step efficiency. More details on the four types as well as sample solutions comparing these types is provided in \cite{maniktala2020jedm}. We used the \textit{global quality} and \textit{absolute progress} measure of step efficiency to define HelpNeed because these metrics led to the strongest (and significant) correlation between students’ training unproductivity and their posttest performance (more details provided in \cite{maniktala2020jedm}). However, each combination of quality and progress captures a different perspective on step efficiency, and was found to significantly predict posttest performance, so we included all four as features for predicting HelpNeed. Next, we elaborate how we updated these quality and progress features to account for students’ hint usage in predicting their HelpNeed.

\section{Method}
We now describe the design of our new HelpNeed and Use (HNU) predictor, how we train it, and the methods we use to evaluate it.

We modified the HN model to incorporate help usage to reduce the number of false negatives -- times when the prior HN model falsely determined that students would not need help, but actually did. We also changed the state representation for reducing the number of false positives -- times when the prior HN model predicted HelpNeed, but students could accomplish the next step without help. We detail these two modifications in the next two sections. 

\subsection{Reducing False Negatives: Incorporating Hint Usage}
While exploring ways to incorporate students’ hint usage in predicting their HelpNeed, we examined the prior work by other researchers and their findings. One common way researchers employed to carry out this task was to include the number of hint requests \cite{feng2009addressing, wang2010representing, wang2011assistance, hawkins2013extending}. However, this was not sufficient for our task. In fact, we included the hint request count as a feature to predict HelpNeed but this feature was weeded out by all the feature selection approaches we applied (more details of the list of features and the feature selection process are available in \cite{maniktala2020jedm}). There are two reasons for this observation. First, in multi-step open-ended domains, a hint request is not equivalent to hint usage -- a student can request a hint and then solve the problem without using it. We have several semesters of data suggesting that students can request hints, not use them, and still complete problems \cite{maniktala2020leveraging, maniktala2020jedm}. So, the number of hint requests is not a reliable feature for predicting HelpNeed. Second, we have six features that record the most recent step’s quality and progress. These features are highly predictive of the HelpNeed in the next step. The predictive power of these features was so high that it overshadowed the small contributions made by not only the hint request features but also the features that counted the number of hints used. So, in order to incorporate the hint usage, we improved the \textit{signal} for the predictor, i.e., the method for computing these quality and progress features for predicting HelpNeed. We also improved the \textit{distribution} used to train the model by employing the HN Adaptive condition’s data (explained in previous section). These modifications are discussed below.

\subsubsection{Improving the Signal by Reducing Gains Estimates by Half}
We improved the signal for HNU predictor by updating the quality and progress features upon \textit{hint usage}. Our method presented here is motivated by the study carried out by Beck et al. on the Help model \cite{beck2008does}. More specifically, they presented knowledge tracing parameters with estimated values for steps both with and without help. They found that the probability of a student already knowing a concept upon carrying out a step decreased to about half its value when help was given. Furthermore, more literature suggests that the probability of a student knowing a concept is lower with help than that without \cite{d2008more}. In our model, this translates to penalizing (reducing) the quality of a problem-solving state, and the step progress upon hint usage, i.e., we want to train the model to interpret an efficient step upon hint usage as having less quality, i.e. demonstrating less knowledge than an efficient step derived without help. 

The next question is: how \textit{much} do we penalize state quality and step progress upon hint usage? To answer this question, we look at the amount of information we provide in a hint for carrying out the next step. Each step consists of two parts: the \textit{justification} and the derived statement. The justification is applying a rule to a set of 1-2 existing nodes, and the derived statement is the result. We hint students on what statement to derive next. So, we provide about half the information needed to carry out the next step. Therefore, we reduce the improvement in state quality and progress by half when a step is completed with hint justification (which is our hint usage metric, see section \ref{s3-hint-use}). Note that we do not reduce a step’s post-state quality and progress by half, but rather, we penalize the \textit{gain} in quality and progress from the previous state. We illustrate this concept in the following example. Consider a step carried out with hint justification. The pre-state of the step has a global quality of 81. The post-state’s original global quality is 87 but because of hint justification, their penalized global quality is 81 + (87 - 81)/2 = 84 instead of 87. This modified credit (absolute progress) for the most recent step  is 10 (84-74) instead of 13 (87-74), since the start state’s global quality is 74. We hypothesize that this reduction in the quality and progress values can enable the tutor to differentiate between efficient steps carried out with and without help.

\subsubsection{Improving the Distribution} 
We also improved the distribution used to train the HNU predictor. We used the HN Adaptive group’s dataset to train the HNU predictor. While we have several dataset with proactive hints given either randomly or using reinforcement learning, the dataset on the HN Adaptive group is the only one that uses a proactive hint policy with a HN predictor, and hence, it is most indicative of student behavior in such an adaptive hint policy.

We performed a 3-fold cross validation on the HN Adaptive dataset (\textbf{3CV-HN-A}) to assess the effectiveness of using this dataset in predicting HelpNeed. Table \ref{table:new-test} shows the result of carrying out this test on Random Forest models for both state-based and state-free predictions. Note that we experimented with classifiers including Random Forest, Decision Tree, Support Vector Classifier, Multi-layer Perceptron, Quadratic Discriminant Analysis, K-Nearest Neighbours, AdaBoost, Naive Bayes, and XGBoost. We also experimented with parameters such as class weights to maximize both recall (proportion of HelpNeed steps correctly predicted) and area under the ROC curve (AUC, the ability of a model to distinguish between HelpNeed and non-HelpNeed steps). We selected Random Forest models because they had the highest recall, i.e., they maximized the chances of predicting HelpNeed when help was needed.  Table \ref{table:new-test} shows the results for this analysis. As in our previous work with state-based and state-free classifiers for HelpNeed, the state-based random forest model has a higher AUC than the state-free random forest model.

\begin{table}
\centering
\begin{tabular}{|c|c|c|c|c|} 
\hline
\multirow{2}{*}{Test} & \multicolumn{2}{l|}{State-based~}                      & \multicolumn{2}{l|}{State-free}                         \\ 
\cline{2-5}
                      & \multicolumn{1}{l|}{Recall} & \multicolumn{1}{l|}{AUC} & \multicolumn{1}{l|}{Recall} & \multicolumn{1}{l|}{AUC}  \\ 
\hline
3CV-HN-A              & .86                        & .83                     & .75                        & .63                      \\ 
\hline
3CV-HN-AC              & .81                        & .80                      & .71                        & .62                      \\ 
\hline
Holdout-HN-A               & .87                        & .80                      & .77                        & .63                      \\ 
\hline
Holdout-HN-AC                & .88                        & .79                     & .79                         & .61                      \\
\hline
\end{tabular}
\caption{Understanding the impact of data used in training the HNU predictor for both state-based and state-free predictions}
\label{table:new-test}
\end{table}

In addition to 3CV-HN-A, we performed three additional tests to understand the impact of training predictive models on different datasets. First, we performed the test \textbf{3CV-HN-AC} where both HN Adaptive and Control (HN-AC) datasets were split into 3 folds (3CV-HN-AC); the models were trained on four folds (two folds per HN Adaptive and Control) and tested on the remaining two folds (one per HN Adaptive and Control). This testing was performed to understand how well a classifier trained on data with and without proactive help would perform in predicting both policies. Similar to 3CV-HN-A, in the 3CV-HN-AC test, the state-based predictor has a better AUC than the state-free predictor. Though 3CV-HN-A and 3CV-HN-AC are not directly comparable because they are tested on different datasets, the higher recall and AUC in 3CV-HN-A suggests that there may be differences in student behavior between the HN Adaptive and Control groups that lower the performance of 3CV-HN-AC models in predicting HelpNeed for HN Adaptive and Control together than it is for 3CV-HN-A models in predicting HelpNeed for the HN Adaptive group alone. 

Next, we performed the test \textbf{Holdout-HN-A}, a holdout testing with training on HN Adaptive and testing on the dataset that was used to train the HN predictor. This testing was performed to assess the generalizability of the model trained on the HN Adaptive dataset in predicting prior datasets with randomly-given proactive hints. Finally, we performed the test \textbf{Holdout-HN-AC}, a holdout test with training on HN Adaptive and Control (HN-AC), and testing on prior datasets. This test was performed to assess how models trained on HN Adaptive are at predicting prior semesters in comparison with models trained on the combination of HN Adaptive and Control.

Table \ref{table:new-test} also shows the results of Holdout-HN-A and Holdout-HN-AC which \textit{can} be compared because they are tested on the same datasets. The results of these tests are similar for both state-based and state-free predictions, with Holdout-HN-A showing a slightly lower recall and slightly higher AUC than Holdout-HN-AC. These results suggest training the models on HN Adaptive alone does not significantly reduce the performance of a classifier in predicting prior semesters’ HelpNeed. 

These holdout test results, in conjunction with 3-fold cross validation on the HN Adaptive group (3CV-HN-A) results, suggest that training the HNU predictor on HN Adaptive group can result in an effective classifier for providing proactive hints, which is also generalizable over prior semesters. Before we present the controlled study that tests this hypothesis, we present another modification we employed in the HNU predictor. Note that, unlike the modification of incorporating hint usage that aims to reduce false negatives, the following modification aims to reduce false positives.

\subsection{Reducing False Positives: Updating State Representation}
\label{sec:s3-usp}
Recall that the HelpNeed predictor comprises the state-based and state-free classifiers (see section \ref{sec:f19_policy}). The state-free predictor is used when a student’s problem-solving state cannot be found in the historical student data. Since the state-free predictor cannot use step efficiency features (quality and progress), we modeled it to have a high recall at the expense of a higher false positive rate. We made this trade-off because we want to correctly predict most HelpNeed steps so that timely intervention is given to students. While we were not able to improve the state-free predictor in this study, we use a method that can reduce the usage of the state-free classifier. Since the state-free predictor is used when a student’s problem-solving state cannot be matched in the historical student data, we aimed to reduce its usage by devising a method to increase state matches with historical data. 

The HelpNeed predictor uses \textit{ordered} state matches, where we take into account the order of statement derivation while finding a matching state in the historical student data. Prior work on the Hint Factory reveals that performing \textit{unordered} state matches with historical data can drastically increase these matches \cite{barnes2011using}. For unordered matches, we do not take into account the order of statement derivation while finding a state matching that of a new student. 

In the next section, we present an experiment carried out to evaluate the HNU predictor that (1) penalizes quality and progress features upon hint usage (improved signal), (2) is trained on the HN Adaptive condition (improved distribution), and (3) uses unordered state representation for finding state-matches with the historical data (to reduce false positives).  

\section{Experiment}
\label{sec:s3_exp}
To understand the impact of incorporating students’ hint usage in the HNU predictor, we conducted a controlled study  where the \textit{HNU Adaptive} condition received proactive hints upon predictions of HelpNeed using the HNU predictor, while the \textit{HNU Control} did not receive proactive hints. Students in both the HNU conditions could request hints on-demand during training. For comparison purposes, this study was setup similar to our prior HN study, where students in the HN Adaptive condition received hints using the original HN predictor, and HN Control, where students did not receive any proactive hints.

\subsection{Hypotheses}
We have the following hypotheses:
 \begin{itemize}
 \item \textbf{Posttest}
 
  \begin{itemize}
 \item \textbf{H1 (efficiency and time)}: Students in the HNU Adaptive condition will form shorter (more efficient) proofs faster in the posttest than those in the HNU Control condition
 \end{itemize}
 
 \item \textbf{Training}
 
 \begin{itemize}
\item \textbf{H2 (false negatives)}: The HNU predictor will have a lower proportion of false negatives (FN) for the HNU Adaptive condition than the HN predictor
\item \textbf{H3 (false positives)}: Using unordered state representation will significantly increase the number of state-matches, thereby, significantly increase calls to the state-based classifier (which has a lower rate of false positives (FP) than the state-free classifier).
\item \textbf{H4 (training productivity)}: Compared to the HNU Control, students in the HNU Adaptive condition will have a better training behavior with fewer HelpNeed steps
\item \textbf{H5 (help usage)} Compared to the HNU Control, students in the HNU Adaptive condition will have a lower possible help avoidance, and higher possible help appropriateness, as measured using the HelpNeed classification and predictor.
\end{itemize} 
 \end{itemize}
 
\subsection{Participants}
The tutor is given as a homework assignment every semester in a discrete math course (CSC 226) at North Carolina State University to undergraduate students in the College of Engineering majoring in Computer Science, Computer Engineering, or Electrical Engineering. The HNU study experiment was conducted with 84 participants in the Spring 2020 semester, and the students were initially given ten days to complete the tutor comprising seven days before the Spring break and three days after it. However, the COVID-19 related lockdown extended the Spring break by a week, so we extended the assignment deadline by a week. We do not have precise demographics for the Spring 2020 HNU study. However, according to the instructor of this course, the demographics for the Spring 2020 HNU study were similar to that of the Fall 2019 HN study; note that the two semesters are within the same Academic Year 2019-2020, and are about 4 months apart. The Fall 2019 College of Engineering demographics included 25.9\% women, 65.9\% white, 9.2\% Asian, 6.2\% Non-resident Alien, 0.3\% American Indian/Native American, 3.4\% Black/African American, 5.7\% Hispanic/Latino, 3.5\% from two or more under-represented minorities, and 5.8\% with unknown race/ethnicity \footnote{Their website does not display demographics for the Spring semesters. More details on Fall 2018, 2019, and 2020 student demographics at NCSU can be found at https://www.engr.ncsu.edu/ir/fast-facts/.}. The CSC 226 course is typically composed of about 60\% sophomores, 30\% juniors, 9\% seniors, and 1\% freshmen. 

\subsection{Procedure}
\label{sec:procedure}
The tutor is divided into four sections: introduction, pretest, training, and posttest. The \textit{introduction} presents two worked examples to familiarize students with the tutor interface. Next, we determine students’ incoming competence in a \textit{pretest} comprising two easy and short problems, solvable with short optimal solution lengths ($\mathit{Mean} = 3.5$, $\mathit{SD} = 0.71$). Stratified random sampling based on pretest performance (see section \ref{sec:pro_perf}) is used to partition 84 HNU study participants into the two conditions, resulting in 42 in \textit{Adaptive} and 42 in \textit{Control}. Next, students go through the \textit{training} section with fifteen problems of varying difficulty. The difficulty of the training problems is between that of the pretest and the posttest, assessed on averaging the optimal solution lengths over all the training problems ($\mathit{Mean} = 4.99$, $\mathit{SD} = 1.32$). Students in both the conditions can request on-demand hints in training but only the Adaptive condition is given proactive hints using the revised HelpNeed predictor -- the HNU predictor. Finally, students take a \textit{posttest} with five problems that are more difficult with longer optimal solution lengths than the other sections ($\mathit{Mean} = 7.25$, $\mathit{SD} = 1.89$). Note that the tutor is designed to provide immediate feedback on rule application errors in all the sections. Among the 84 participants, 74 (36 in Adaptive and 38 in Control) completed the tutor and one Control group participant’s data was removed  because of log errors. We performed a $\chi^2$ test of independence to examine the impact of completion rate and system errors on the groups and found no significant differences among the two groups: $\mathit{\chi^2}(2, \mathit{N} = 84) = .03,  \mathit{p} = .99$. This implies that the group sizes were not significantly impacted by the tutor completion rate or logging errors. Finally, we analyzed the HNU study data for 73 participants (36 in Adaptive and 37 in Control).

\subsection{Performance Metrics}
\label{sec:pro_perf}
We measure performance using three metrics: solution length, problem-solving time, and rule application accuracy \cite{maniktala2020leveraging, maniktala2020jedm}. In open-ended well-structured domains such as logic, forming shorter (more efficient) proofs, taking less time, and making fewer mistakes reflects more expert-like problem-solving. We now define each of the three performance metrics. \textit{Length} of a solution is the number of statements derived in that solution. To calculate the length of a tutor section (pretest, training, or posttest), we sum the length of solutions to all problems solved in that section. We measure length only for the problems that are complete. Since we analyze the dataset for students who completed the tutoring assignment, i.e., they completed all the problems, so we can use this metric to compare the conditions. Next, similar to other studies \cite{kardan2015providing,tchetagni2002hierarchical}, we assess students on their problem-solving \textit{time}. We cap each click-based action time to five minutes (to reduce noise)\footnote{The $3^{rd}$ quartile of action time in the HNU study is 4.1s, and only 2721 out of 214575 actions had an action time greater than five minutes} and sum this capped action time over a section to calculate the time metric. Finally, the performance metric of \textit{accuracy} is defined as the number of correct rule applications divided by the total number of rule applications. 

To check that the data met assumptions for t-tests, we used the Shapiro-Wilk’s test and Levene’s test, as well as visually inspecting the data via Q-Q plots. We used Welch’s t-test for distributions that passed the Shapiro Wilk’s test but not the Levenes’ test. Data that did not meet the assumptions were transformed using log or square-root transformations, then reinspected. For data that still did not meet assumptions, either Mann-Whitney U test or Kruskal-Wallis test with Dunn post hoc test and Bonferroni corrections were used. For clarity, all data in tables are reported before transformation.

\subsection{Hint Usage}
\label{s3-hint-use}
As mentioned earlier, we use hint justification as the measure of hint usage. The process of hint justification involves selecting existing statements and rules to derive the hinted statement. Hint justification rate (HJR) is defined as the proportion of hints given (on-demand or proactive) that are correctly justified. We also investigate student help behaviors such as possible help avoidance, abuse, and appropriateness. We defined them in our prior HelpNeed study \cite{maniktala2020jedm}, and are provided in Table \ref{tab:s3-haaa-def}. Note that we add a prefix \textit{possible} to these behaviors because HelpNeed does not represent ground truth as classified by experts. Rather, HelpNeed is our classification of steps needing help, and the prediction is a heuristic measure.

\begin{table}
\centering
\caption{Definition of possible help avoidance, abuse, and appropriateness using the HelpNeed classification, described in our prior work \cite{maniktala2020jedm}}
\label{tab:s3-haaa-def}
\begin{tabular}{|l|l|} 
\hline
Help Behavior Metric                          & Definition                                                                                 \\ 
\hline
Possible
  Help Avoidance       & \begin{tabular}[c]{@{}l@{}}\%steps with observed HelpNeed \\but no hints were 
  requested or received\end{tabular}    \\ 
\hline
Possible
  Help Abuse           & \begin{tabular}[c]{@{}l@{}}\%steps with no
  predicted or observed HelpNeed \\but hints 
  were requested\end{tabular}    \\ 
\hline
Possible
  Help Appropriateness & \begin{tabular}[c]{@{}l@{}}\%steps with
  predicted HelpNeed \\and hints were received\end{tabular}                    \\
\hline
\end{tabular}
\end{table}
\section{Results}

In this section, we present the analyses investigating our five hypotheses: H1 on improved posttest performance, H2 on fewer false negatives, H3 on fewer false positive, H4 reduced training HelpNeed, and H5 on improved help usage.

\subsection{Performance Metrics}
This section provides an overview of student performance in the two HNU conditions \{Adaptive, Control\} using the metrics \{Length, Time, Accuracy\} discussed in section \ref{sec:pro_perf}. We first investigate hypothesis H1 on whether the HNU Adaptive students have better posttest length and time than their Control peers. We also briefly contrast this comparison with that of the HN study. 

Table \ref{tab:s3_performance} shows the distribution parameters of students’ pretest, training, and posttest performance in the two HNU Conditions \{Adaptive, Control\}.  As expected, no significant differences were found between the two conditions in the pretest performance metric of (1) \textit{length}: \textit{U} = 659, \textit{p} = .47, (2) \textit{time}: \textit{U} = 649, \textit{p} = .43, and (3) \textit{rule application accuracy}: \textit{t}(72) = 0.30, \textit{p} = .77. This confirms that our stratified random sampling assignment balanced HNU Adaptive vs. Control conditions' incoming competence (see section \ref{sec:procedure}).

\begin{table}
\caption{Distribution parameters for the performance metrics of the two HNU conditions in the pretest, training, and posttest section of the tutor}
\centering
\begin{tabular}{|c|r|c|c|c|} 
\hline
Section                   & Metric   & Adaptive  & Control    & \multicolumn{1}{c|}{\textit{p}}  \\ 
\hline
\multirow{3}{*}{Pretest}  & Length (\#nodes)  & 16 (5)    & 16 (5)     & .47                     \\ 
\cline{2-5}
                          & Time  (min)    & 54 (49)   & 57 (46)    & .43                     \\ 
\cline{2-5}
                          & Accuracy & .41 (.16) & .45 (.2)   & .77                     \\ 
\bottomrule
\color{blue}{}\multirow{3}{*}{Training} & \color{blue}Length (\#nodes)   & \color{blue}86 (9)    & \color{blue}92 (15)    & \color{blue}.04*                     \\ 
\cline{2-5}
                          & \color{blue}Time (min)    &\color{blue} 64 (27)   &\color{blue} 90 (43)    &\color{blue} .01*                     \\ 
\cline{2-5}
                          & \color{blue}Accuracy & \color{blue}.74 (.10) &\color{blue} .69 (.13) & \color{blue}.05*                     \\ 
\bottomrule
\multirow{3}{*}{Posttest} & \color{blue}Length (\#nodes)   &\color{blue} 39 (8)    & \color{blue}\color{blue}44 (12)    & \color{blue}.06                     \\ 
\cline{2-5}
                          & Time  (min)    & 31 (30)   & 30 (27)    & .98                     \\ 
\cline{2-5}
                          &\color{blue} Accuracy & \color{blue}.76 (.11) &\color{blue} .71 (.11)  &\color{blue} .04*                     \\
\hline
\end{tabular}
\label{tab:s3_performance}
\end{table}

Next we examine the differences in the \textit{training} performance between the HNU Adaptive and Control conditions. Significant differences were found between the two conditions in the training (1) \textit{length}: \textit{t}(72) = 1.75, \textit{p} = .04, Cohen’s d = .49, (2) \textit{time}: \textit{t(72)} = 2.25, \textit{p} = .01, Cohen’s d = .72, and (3) \textit{accuracy}: \textit{t}(72) = 1.65, \textit{p} = .05, Cohen’s d = .43. This is a marked difference from the HN study, where the HN Adaptive condition was only marginally significantly better in solution length than the HN Control, with no significant differences in the training time or accuracy. This provides evidence to support that the HNU predictor improved the training performance for the HNU Adaptive condition. We further investigate students’ training behavior within the HNU study as well as contrast the differences in the Adaptive and Control conditions between HNU and HN studies in section \ref{s3:train}.

Next, a Mann-Whitney U test on posttest length shows a marginally significant difference between the two HNU conditions (\textit{U} = 495, \textit{p} = .06) with the Adaptive condition (\textit{Mean} = 39, \textit{SD} = 8) forming marginally significantly shorter proofs than the Control (\textit{Mean} = 44, \textit{SD} = 12), and a moderate effect size (Cohen's d = .49). Next, on the transformed posttest time, no significant differences (\textit{t}(72) = .03, \textit{p} = .98) were found between the two conditions, with students in the Adaptive condition (\textit{Mean} = 31min, \textit{SD} = 30min) spending a similar amount of time on the posttest as the Control (\textit{Mean} = 30min, \textit{SD} = 27min). These observations partly confirm our H1 hypothesis. Interestingly, we found significant differences in the posttest accuracy between the two HNU conditions (\textit{t}(72) = 1.75, \textit{p} = .04), with the Adaptive condition (\textit{Mean} = .76, \textit{SD} = .11) having significantly higher accuracy than the Control (\textit{Mean} = .71, \textit{SD} = .11), with a moderate effect size (Cohen’s d = .45). We hypothesize that this result may be a consequence of effectively predicting \emph{when} students need help so that they get more practice with the targeted rules and less distraction from rules that are present on the screen but are rarely used in efficient solutions. We further discuss this improved accuracy in section \ref{sec:disc}.

Note that in the HN study, students in the HN Adaptive condition formed significantly shorter proofs in the posttest in significantly less time (11 minutes faster on average) than the Control. The HNU predictor led the HNU Adaptive condition to save 26 minutes in training compared to the HNU Control. We further investigated students’ training behavior in the next section to understand the impact of the revisions made in the HNU predictor and how it may have impacted the posttest results.

\subsection{Training} \label{s3:train}
In this section, we investigate whether incorporating students’ help usage in the HNU model led to improved HelpNeed predictions during training (H2 and H3), students’ training unproductivity (H4), as well as their help-usage behaviors (H5) such as help avoidance, help appropriateness, and help abuse. 

\subsubsection{Did we improve the predictor?} \label{sec:s3-didwe}
In this section, we investigate hypothesis H2 (false negatives) and H3 (false positives), i.e., whether incorporating students’ help usage in our model leads to improved HelpNeed predictions during the tutor's training section. H2 states that the HNU predictor has a lower proportion of false negatives (FN) for the HNU Adaptive condition participants than the HN predictor. Note, we don’t compare the false negatives observed in the HN study with that of HNU, but rather, compare both predictors on the HNU Adaptive condition’s data. By using the same data, we can control for extraneous factors. We also don’t compare the two predictors on the HN Adaptive condition’s data because the HNU predictor was trained on that data. Hence, comparing the two predictors on the HNU Adaptive condition ensures an honest assessment.

\begin{figure*}
\centering
\includegraphics[width=0.45\columnwidth]{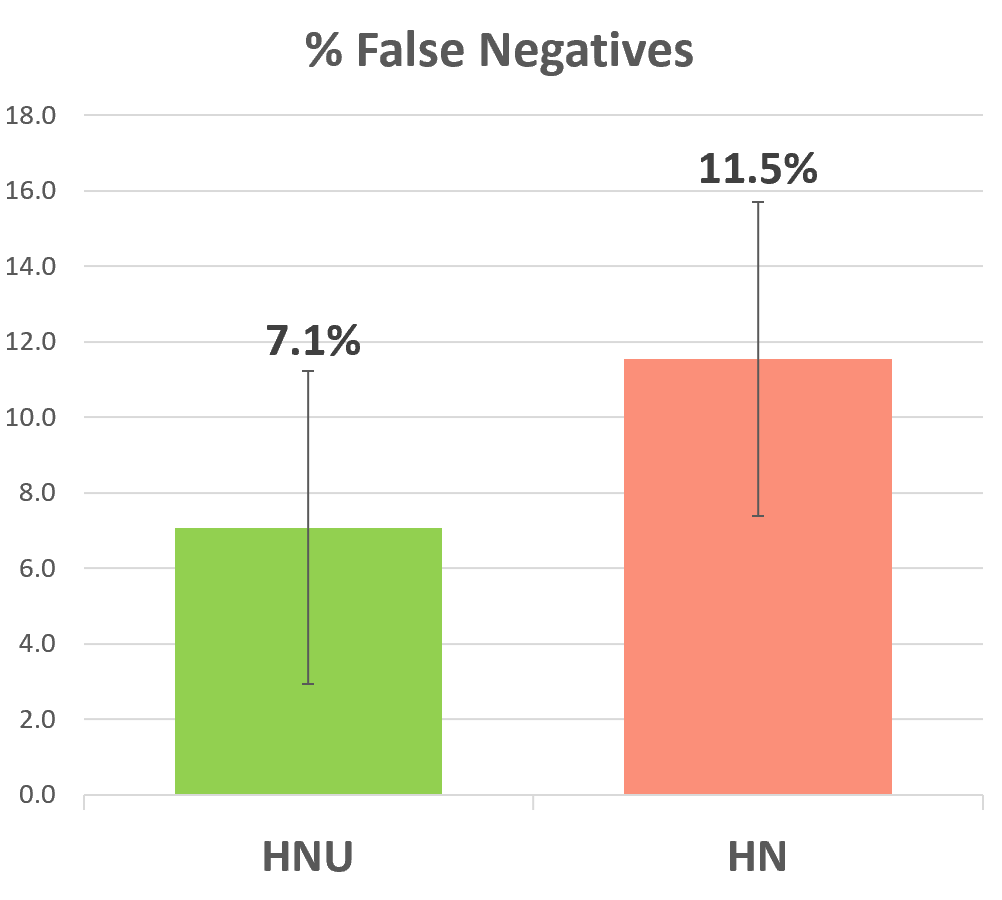}
\caption{The HNU predictor has significantly lower  percentage of false negatives on the HNU Adaptive condition data when compared to the HN predictor (\textit{p} = .002)}
\label{fig:fnr}
\end{figure*}

Figure \ref{fig:fnr} shows that, on average, the HNU predictor has 7.1\% instances with a false negative (SD = 4.2), and the HN predictor has 11.5\% false negatives (SD = 9.2). A Mann Whitney test shows a significant difference (\textit{U} = 768.5, \textit{z} = 2.93, \textit{p} = .002) between the two predictors in the percentage of false negatives, with the HNU predictor performing significantly better than the HN predictor. This confirms our H2 hypothesis.

Next, we investigate the impact of unordered state representation on the HNU predictor. Recall, we hypothesized (H3) that unordered state representation will increase the number of state matches which will, in turn, increase the calls made to the state-based classifier (which has a lower rate of False Positives (FP) than the state-free classifier). The HN study used ordered state-matches leading to 70.6\% calls to the state-based predictor. Whereas, in the HNU study, we were able to attain 90.3\% state-based calls because of the unordered state representation. A chi-squared test shows a significant difference in the proportion of calls to the state-based vs. state-free classifiers between the two studies ($\mathit{\chi^2}(1, \mathit{N} = 24,311) = 32.82,  \mathit{p}  < .001$). This confirms our H3 hypothesis. The reduction in the number of state-free calls because of the unordered state representation is noteworthy because the state-free classifier errs on the side of higher recall at the expense of more false negatives, i.e., it errs on the side of providing more proactive hints to ensure a high recall. So, increasing the number of state-based calls ensures that we reduce the number of unnecessary proactive hints given to students. We discuss the number of hints given to students (proactive or on-demand) in section \ref{sec:hints_given_used}.

\subsubsection{Did the improvement in the predictor reduce training unproductivity?}
\label{sec:unp_s3}
In the HN study, even though the Adaptive condition had significantly fewer Far Off steps than the Control, there were no significant differences in the total number of HelpNeed training steps between the HN Adaptive and Control conditions \cite{maniktala2020jedm}. We hypothesized that HNU’s improved predictions will lead the HNU Adaptive condition to have significantly fewer HelpNeed training steps than the HNU Control (H4). We found a significant difference  (\textit{p} = .02)  in the HelpNeed steps between the HNU Adaptive condition (\textit{Mean} = 25, \textit{SD} = 18) and the HNU Control (\textit{Mean} = 43, \textit{SD} = 49). This suggests that the HNU predictor successfully predicted and prevented training unproductivity (HelpNeed steps) for the HNU Adaptive condition, confirming hypothesis H4. 


\begin{table}
\caption{Distribution parameters for students’ training steps in the two HNU conditions. T-tests show significant differences between the two conditions in the Expert (\textit{p} = .02), Far Off (\textit{p} = .04), and Futile (\textit{p} = .02) steps.}
\centering
\begin{tabular}{|l|l|l|l|l|r|} 
\hline
\multicolumn{2}{|c|}{Step Behavior }     & Description                      & Adaptive & Control & \multicolumn{1}{c|}{\textit{p}}  \\ 
\hline
\multirow{3}{*}{non-HelpNeed} &\color{blue}Expert
      &  \color{blue}quick
  efficient steps              & \color{blue}82 (17)  & \color{blue}74 (16)&\color{blue} .02*                    \\ \cline{2-6} 
&Strategic
   & long efficient steps                 & 21 (7)   & 22 (10) & .23                     \\ \cline{2-6} 
&Opportunistic      & singular quick
  inefficient steps           & 8~ (5)   & 8 (5)   & .38                     \\ 
\hline
\multirow{2}{*}{HelpNeed} & \color{blue}Far
  Off   & \color{blue}consec.
  quick inefficient steps & \color{blue}15 (13)  & \color{blue}26 (34) &\color{blue} .04*                     \\ \cline{2-6} 
&\color{blue}Futile
     & \color{blue}long
  inefficient steps             & \color{blue}10 (7)  & \color{blue}17 (18) & \color{blue}.02*                     \\ \bottomrule
  
 \multicolumn{3}{|r|}{Total Training Steps}                                                                        & 136 (31)                       & 148 (52)                       & .08                         \\ \bottomrule
\hline
\end{tabular}
\label{tab:training_steps_lower}
\end{table}

We now drill down on the step behavior and compare the number of each training step-type between HNU Adaptive and Control conditions, as shown in Table \ref{tab:training_steps_lower}. The HNU Adaptive group has significantly more Expert steps than the HNU Control (Adaptive: 82, Control: 74, \textit{p} = .02). Further, the HNU Adaptive condition not only has significantly fewer Far Off (Adaptive: 15, Control: 26, \textit{p} = .04) but also has significantly fewer Futile steps than the Control (Adaptive: 10, Control: 17, \textit{p} = .02). Finally, on the total training steps, a t-test shows that the HNU Adaptive condition took marginally significantly fewer total training steps on average compared to the HNU Control condition: \textit{t}(72) = 1.4, \textit{p} = .08. It is interesting that the HNU Adaptive condition has marginally significantly fewer total steps even when they have significantly more Expert steps. This is because when students follow expert-like strategies, they can solve the problems more efficiently with fewer steps overall. The significantly higher Far Off and Futile steps in the HNU Control condition may be a result of help avoidance because students may not know when to seek help \cite{pena2011improving,azevedo2004does}. More details on student help avoidance are discussed in section \ref{sec:help-seek}. 

\subsubsection{How many hints were given and used?}
\label{sec:hints_given_used}
In this section, we further investigate the sources of differences in the training behavior between the HNU Adaptive and Control conditions. Table \ref{tab:s3-hint_usage} shows the mean and standard deviation of the total number of proactive, on-demand, and overall hints received by students in the two HNU conditions across training problems (the top part), and their hint justification rate (HJR, the bottom part). Note that the HNU Control condition was not provided with proactive hints, and thus only their on-demand hints count toward their total hints. 

\begin{table}[]
\centering
\caption{Distribution parameters for the number of hints given and the hint justification rate in the two HNU Adaptive and Control conditions}
\begin{tabular}{cc|c|c|c|}
\cline{3-5}
\multicolumn{1}{l}{}                                                                                          & \multicolumn{1}{l|}{} & Adaptive       & Control                  &                              \textit{p} \\ \hline
\multicolumn{1}{|c|}{}                                                                                        & Proactive            & 27 (5)  & - &-     \\ \cline{2-5} 
\multicolumn{1}{|c|}{}                                                                                        & On-demand             & 10 (11)   & 9 (11)            &  .28              \\ \cline{2-5} 
\multicolumn{1}{|c|}{\multirow{-3}{*}{\begin{tabular}[c]{@{}c@{}} \# Hints \\ Received\end{tabular}}}             & Total Hints                & 37 (12) & 9 (11)            & \textless{}.01*              \\ \hline
\multicolumn{1}{|c|}{}                                                                                        & Proactive            & 93\% (6\%)  & - & -     \\ \cline{2-5} 
\multicolumn{1}{|c|}{}                                                                                        & On-demand             & 96\% (8\%) & 86\% (25\%)            & .02*                        \\ \cline{2-5} 
\multicolumn{1}{|c|}{\multirow{-3}{*}{\begin{tabular}[c]{@{}c@{}}\% Hint \\ Justification \\ Rate (HJR) \end{tabular}}} & Total HJR                & 93\% (6\%)  & 86\% (25\%)            & .44                        \\ \hline
\end{tabular}
\label{tab:s3-hint_usage}
\end{table}

Students in the HNU Adaptive condition on average received 27 proactive hints (19.6\% of steps). This is similar to the average number of proactive hints received by the HN Adaptive Condition (\textit{Mean} = 28; 22.8\% of steps). This is particularly interesting because even though the two predictors lead to a similar proactive hint count, only the HNU predictor significantly reduced the unproductive HelpNeed steps (see \ref{sec:unp_s3}). We believe this is a result of reducing both the false negatives (and thereby increasing the chances of students receiving hints when they are needed), and false positives (and thereby reducing the chances of providing unnecessary help).

Next, we observed no significant differences in the number of on-demand hint between the two HNU conditions (\textit{U} = 583.5, \textit{z} = 1,1, \textit{p} = .28), with HNU Adaptive condition having \textit{Mean} = 10 (\textit{SD} = 11), and Control having \textit{Mean} = 9 (\textit{SD} = 11). Next, a Mann Whitney U test on the total hints for the two HNU conditions shows a significant difference (\textit{U} = 82, \textit{z} = 6.5, \textit{p} $<$ .01). While one can argue that the increased number of total hints could have improved the HNU Adaptive condition’s posttest performance, our prior study \cite{maniktala2020jedm} suggested that simply receiving more proactive hints at random times can be harmful, so it is important to identify when help is needed. Ideally, we would compare the HNU policy with a policy that provides the same number of proactive hints randomly. However, this is difficult to achieve, since the policy is adapting to individual students and the number of proactive hints per student is neither predetermined nor consistent. 

For the HJR in the HNU study, the total hints HJRs are high for both conditions (Adaptive: \textit{Mean} = 93\%, \textit{SD} = 6\%; Control: \textit{Mean} = 86\%, \textit{SD} = 25\%), and no significant differences were found between the two conditions (\textit{U} = 562.5, \textit{z} = 0.16, \textit{p} = .44), which suggests that our hints are well-accepted by students in both the HNU conditions. The HNU Adaptive condition justified most of their proactive hints (mean proactive HJR =  93\%). This affirms our prior results that students in the Adaptive condition find it easy to incorporate the unsolicited hints in their solutions \cite{maniktala2020leveraging}. Interestingly, unlike the HN study, we observed a significantly higher on-demand HJR for the HNU Adaptive condition than the HNU Control (HNU \textit{means}: Adaptive: 96\%, and Control: 86\%, \textit{U} = 375.5, \textit{z} = 2.0, \textit{p} =  .02). Further, compared to HN study, students in HNU study had a higher on-demand HJR for the Adaptive condition (HN  Adaptive: 87\%, HNU Adaptive: 96\%) and a lower on-demand HJR for the Control condition (HN Control: 90\%, HNU Control: 86\%). We further looked into students’ on-demand hint usage to understand what may have caused this behavior. We found that on average, the HNU Adaptive group received an on-demand hint to derive the conclusion in 39\% of their on-demand hints, with an SD = 40\%, and max = 95\% for an on-demand hint count = 21. Since students requested these hints, it suggests that, in these situations, the Adaptive group was not aware that they could derive the conclusion in the next step. On the other hand, not a single student in the HNU Control group received an on-demand hint to derive the conclusion. This may suggest that some Adaptive group students may have gamed the on-demand hints. We discuss it further in the next section. 

\subsubsection{Did we reduce help avoidance and increase help appropriateness?}
\label{sec:help-seek}
In this section, we investigate H5 that students in the HNU Adaptive condition will have lower possible help avoidance, and higher possible appropriate help, as measured using the HelpNeed classification and predictor, when compared to the Control. We compare possible help abuse as well to determine whether the Adaptive hints impacted gaming behaviors, where students game or overuse hints to complete the problems faster without learning. We also compared the HNU conditions with that of HN to understand the impact of incorporating help usage into the HNU predictor.

\begin{figure*}
\centering
\caption{Comparing students' help behavior across the four conditions: HN Adaptive, HN Control, HNU Adaptive, and HNU Control }
\begin{subfigure}{.45\textwidth}
  \centering
  \caption{HNU Adaptive condition has significantly lower possible help avoidance than the other three conditions}
  \includegraphics[width=\linewidth]{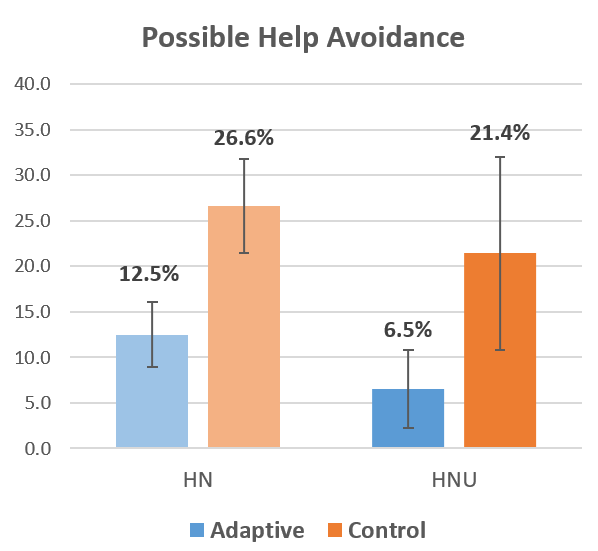}
  
  \label{fig:avoid}
\end{subfigure}%
\hspace{2em}
\begin{subfigure}{.45\textwidth}
  \centering
  \caption{HNU Adaptive condition has significantly higher possible help appropriateness than the other three conditions}
  \includegraphics[width=\linewidth]{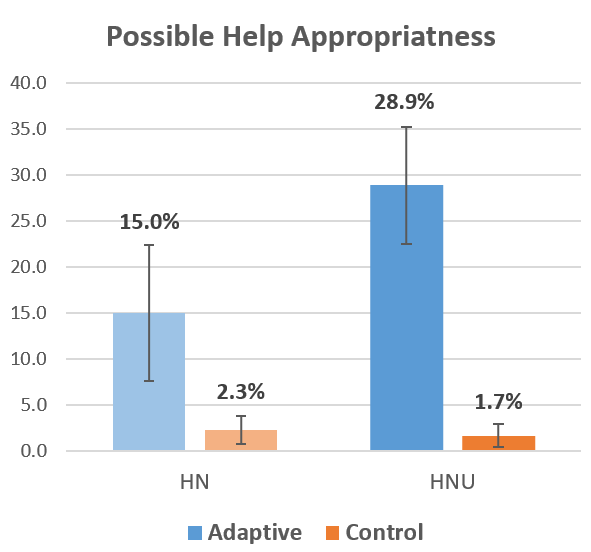}
  
  \label{fig:apt}
\end{subfigure}

\vspace{2em}
\begin{subfigure}{.45\textwidth}
  \centering
  \caption{HNU Adaptive condition has significantly high possible help abuse than the HN Adaptive condition}
  \includegraphics[width=\linewidth]{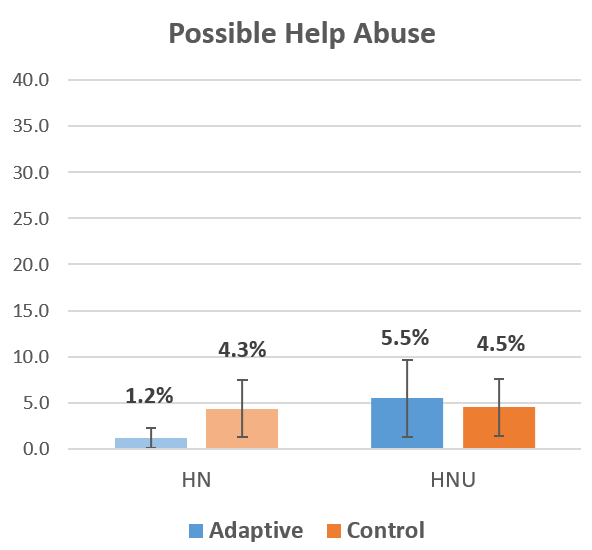}
 
  \label{fig:abuse}
\end{subfigure}
\label{fig:s3-helpseeking}
\end{figure*}

Figure \ref{fig:s3-helpseeking} shows a comparison in possible avoidance, abuse, and appropriateness between the four conditions. We used Kruskal-Wallis test with Dunn post hoc test and Bonferroni corrections. We first discuss possible \textit{help avoidance} shown in Figure \ref{fig:avoid}. A significant difference (\textit{H}(3, N = 184) = 30.3, \textit{p} $<$ .001) was found between the four conditions. The Dunn post hoc test with Bonferroni corrections shows a significant or marginally significant difference between all pairs except HNU and HN control conditions (\textit{p} = 1.0). The HNU Adaptive condition (\textit{Mean} = 6.5\%, \textit{SD} = 4.3\%) has significantly less possible help avoidance than all other conditions: HNU Control (\textit{Mean} = 21.4\%, \textit{SD} = 4.3\%, \textit{p} $<$ .001), HN Control (\textit{Mean} = 26.6\%, \textit{SD} = 5.2\%, \textit{p} $<$ .001), and the HN Adaptive condition (\textit{Mean} = 12.5\%, \textit{SD} = 3.5\%, \textit{p} = .0). 

Next, Figure \ref{fig:apt} shows possible \textit{help appropriateness}, i.e., the proportion of steps where HelpNeed was predicted, and help was either given or requested. A Kruskal-Wallis test shows a significant difference in possible help appropriateness between the four conditions (\textit{H}(3, N = 184) = 135.9, \textit{p} $<$ .001). The Dunn post hoc test with Bonferroni corrections shows a significant difference between all pairs except HNU and HN control conditions (\textit{p} = 1.0). The HNU Adaptive condition (\textit{Mean} = 28.9\%, \textit{SD} = 6.4\%) has significantly (\textit{p} $<$ .001) higher possible help appropriateness than all other conditions: HNU Control (\textit{Mean} = 1.7\%, \textit{SD} = 1.3\%), HN Control (\textit{Mean} = 2.3\%, \textit{SD} = 1.5\%), and the HN Adaptive condition (\textit{Mean} = 15.0\%, \textit{SD} = 7.4\%). These results confirm our H5 hypothesis that students in the HNU Adaptive condition will have lower possible help avoidance, and higher possible appropriate help compared to the Control. It is interesting to note that while we only hypothesize significant differences between HNU Adaptive and Control conditions, the HNU Adaptive condition had significantly lower possible help avoidance and higher possible appropriate help than the HN Adaptive condition as well.

Finally, we look into possible \textit{help abuse} shown in Figure \ref{fig:abuse}. A Kruskal-Wallis test shows a significant difference in possible help abuse between the four conditions (\textit{H}(3, N = 184) = 8.6, \textit{p} = .03). Next, we applied the Dunn post hoc test with Bonferroni corrections. Interestingly, we found that the HNU Adaptive condition (\textit{Mean} = 5.5\%, \textit{SD} = 4.2\%) has a significantly (\textit{p} = .02) higher possible help abuse than the HN Adaptive condition (\textit{Mean} = 1.2\%, \textit{SD} = 1.1\%) . Further, we found no significant differences in the possible help abuse between the HNU Adaptive and Control conditions (\textit{p} = 1.0). It is possible that COVID-19 lockdown may have interacted with HNU Adaptive students’ dependence on hints. We further discuss this in the next section. We also hypothesize that the increased possible help abuse in the HNU Adaptive condition may have negatively impacted their posttest time. 

\section{Discussion} \label{sec:disc}
\subsection{Adaptive Hint Policy: HN vs. HNU}
In this section, we discuss the results of providing partially worked steps as proactive hints using our HNU predictor (HelpNeed predictor that takes into account students help usage), and our post-hoc comparison between the HN (without incorporating help usage) and HNU (with incorporating help usage) studies for solving the assistance dilemma in Deep Thought. 

A direct consequence of incorporating help usage in predicting HelpNeed is the significantly lower proportion of false negatives. This is important because it increases the chances of students receiving help when it is needed. A consequence of this reduction in false negatives is that the HNU Adaptive hint policy successfully delivered proactive hints more appropriately than the HN Adaptive hint policy. While the HN predictor significantly reduced Opportunistic and Far Off steps in training for the HN Adaptive condition when compared to the HN Control, their total training HelpNeed steps did not significantly reduce. However, unlike the HN study, the HNU predictor did lead students in the HNU Adaptive condition to have significantly fewer training HelpNeed steps than their Control peers. Therefore, incorporating students’ help usage in modeling student behavior effectively reduced their training unproductivity. 

A result of the reduced training unproductivity is that the HNU Adaptive condition performed significantly better on the training performance metrics of time, length, and accuracy than the HNU Control condition. Similar to the HN study, students in the HNU Adaptive condition effectively learned efficient  strategies in training and formed shorter solutions in the posttest than their Control peers. But unlike the HN study, the HNU Adaptive condition was not significantly faster than the Control. However, overall, the HNU Adaptive policy helped students save more time (about half an hour) on the tutor than the HN Adaptive policy (about eleven minutes). We suspect that the increased possible help abuse in the HNU Adaptive condition could have negatively impacted their overall posttest time. 

The HNU Adaptive condition received proactive hints on average in only 19.6\% of training  steps (see section \ref{sec:hints_given_used}), and yet it led to a significantly lower possible help avoidance and significantly higher possible help appropriateness when compared to the HNU control and the HN Adaptive conditions. This suggests that the revisions in the HNU predictor helped ensure tutor assistance was administered more appropriately. However, students in the HNU Adaptive condition requested more on-demand hints and had a significantly higher possible help abuse than the HN Adaptive condition (Mean Possible Help Abuse - HN Adaptive: 1.3\% steps, HNU Adaptive: 5.5\% steps). It is possible that receiving proactive hints when they were needed increased student reliance on hints.  It is also possible that the COVID-19 related lockdown contributed towards increased stress among students \cite{son2020effects, bono2020stress}, which may have led the Adaptive condition students (who were more prone to train with hints as reflected in their higher HJR) to game the on-demand hints more often than they otherwise would. Several studies have suggested ways to deal with such gaming behavior. For example, one strategy is to introduce mandatory delays before a student could request a hint \cite{aleven2001helping, murray2005effects}. 

\subsection{HNU Predictor and Targeted Rule Practice}
Next, we investigate our post-hoc hypothesis that the HNU Adaptive condition formed shorter solutions than their Control peers because they received more practice on the \textit{targeted rules}. Targeted rules per problem are the ones that lead to the most efficient (shortest) solutions. Since students are provided a list of about twenty rule options, one important aspect of learning to solve logic proofs is to differentiate between applicable rules that will never lead to a solution, rules that lead to inefficient solutions, and those that lead to efficient solutions. We hypothesized that more practice with the targeted rules helped students in the HNU Adaptive condition to build effective proof solving strategies. More practice with targeted rules would also explain the unexpected consequence of the HNU Adaptive condition having a significantly higher posttest accuracy than the Control. We tested this hypothesis by comparing the number of correct and incorrect rule applications on the targeted rules in the training section between the two HNU conditions.

\begin{figure*}
\centering
\caption{The HNU Adaptive has significantly fewer \textit{incorrect targeted rule applications} (\textit{p} = .02), and significantly more \textit{correct targeted rule applications} (\textit{p} = .02) than HNU Control. And, there are no significant differences in \textit{non-targeted correct and incorrect rule applications} between the two groups}
\begin{subfigure}{.45\textwidth}
  \centering
  \caption{Incorrect Targeted Rule Application}
  \includegraphics[width=\linewidth]{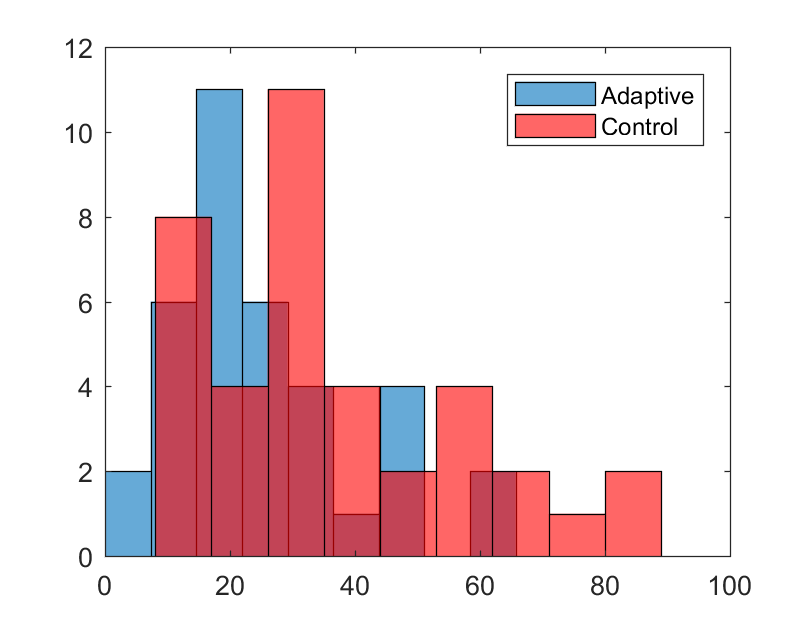}
  
  \label{fig:incorrect-target}
\end{subfigure}%
\hspace{2em}
\begin{subfigure}{.45\textwidth}
  \centering
  \caption{Correct Targeted Rule Application}
  \includegraphics[width=\linewidth]{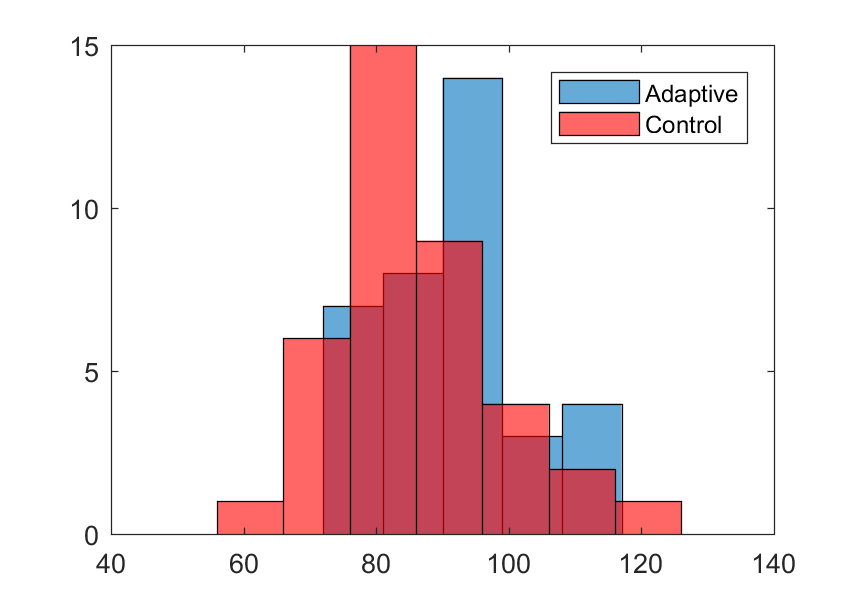}s
  
  \label{fig:correct-target}
\end{subfigure}

\vspace{2em}
\begin{subfigure}{0.75\textwidth}
  \centering
  \caption{Non-targeted Rule Application}
  \includegraphics[width=\linewidth]{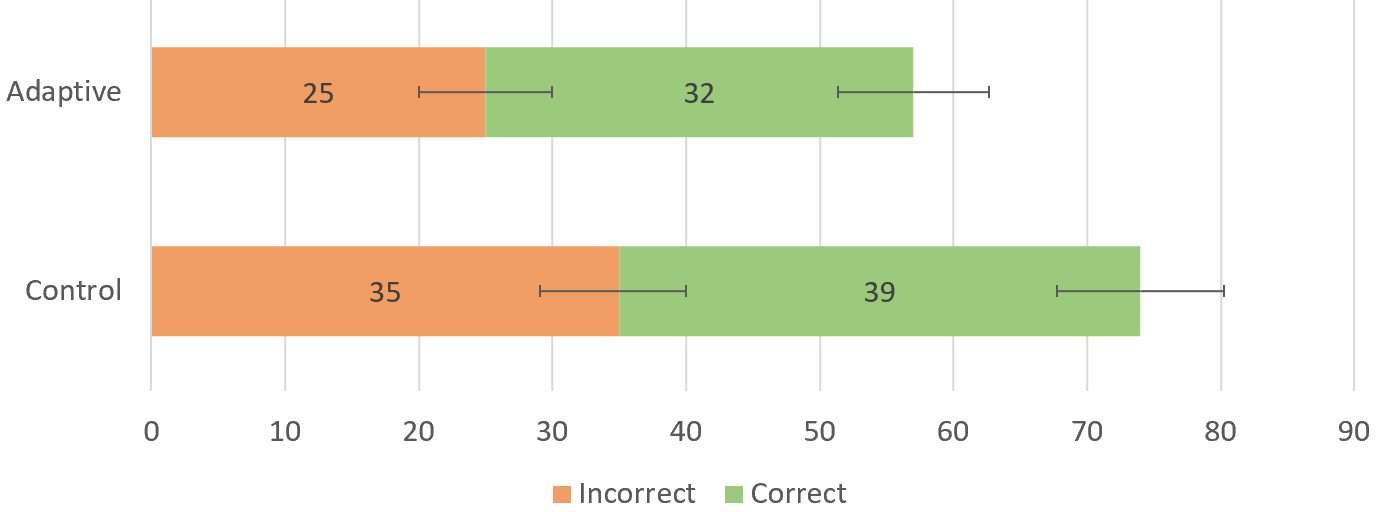}
 
  \label{fig:non-tagerted}
\end{subfigure}
\label{fig:eval-posthoc-hyp}
\end{figure*}

When we compare the total targeted rule applications (correct + incorrect) between the two HNU conditions, we see no significant differences: Adaptive with \textit{Mean} = 117 (\textit{SD} = 21), Control with \textit{Mean} = 123 (\textit{SD} = 30), \textit{U} = 609, \textit{p} = .42. But, interestingly, we found a significant difference in the number of both incorrect and correct targeted rule applications between the two HNU conditions with the HNU Adaptive condition performing better, as shown in Figure 3a and 3b respectively. On the correct applications of targeted rules, the HNU Adaptive (\textit{Mean} = 91, \textit{SD} = 11) had significantly higher application (\textit{U} = 485, \textit{p} = .02) than the HNU Control (\textit{Mean} = 85, \textit{SD} = 12). Furthermore, on the incorrect application of targeted rules, the HNU Adaptive (\textit{Mean} = 26, \textit{SD} = 15) had significantly lower rule applications (\textit{U} = 490, \textit{p} = .02) than the HNU Control (\textit{Mean} = 38, \textit{SD} = 26). These observations are interesting because we provide all students with a list of targeted rules in each training problem. It seems that both groups tried to use targeted rules in training problems, but the HNU Adaptive group was more effective in applying them.  The HNU Adaptive condition not only has significantly more correct application of targeted rules during training than the HNU Control condition, they also have significantly fewer incorrect applications of targeted rules. We suspect that hinting students on \textit{what to derive next} using the HNU predictor allows them to limit their search to just the problem’s targeted rules. 

Next, on the total non-targeted rules, we observe no significant differences in total (correct + incorrect) application between the two HNU conditions (\textit{U} = 640, \textit{p} = .64), with the HNU Adaptive condition (\textit{Mean} = 57, \textit{SD} = 32) having fewer, albeit insignificantly, such applications than the HNU Control (\textit{Mean} = 74, \textit{SD} = 75). Figure 3c compares the correct and incorrect non-targeted rule applications individually, we again see no significant differences between the two HNU conditions (correct: \textit{p} = .28, incorrect: \textit{p} = .32). The slightly higher correct and incorrect non-targeted rule applications for the HNU Control condition suggests that they were more likely to attempt and/or follow non-optimal solution paths during training.

The students in the HNU Adaptive group formed significantly shorter proofs in training with significantly higher accuracy than the Control. Together with the data on targeted and non-targeted rule applications, this implies that the HNU Adaptive group was more likely to correctly apply targeted rules when they were needed for more efficient proofs in training than the Control, and it confirms this post-hoc hypothesis.

\subsection{Impact of Improved Signal on the HNU Predictor}
The HN Adaptive condition’s data was most indicative of students’ help usage under an adaptive hint policy that used a HelpNeed predictor. So, we knew training the HNU predictor on this data would make the predictor more effective. However, is it necessary to collect and apply ML to the HN Adaptive condition’s data to form an effective HNU Adaptive policy? Answering this question is important because, as an ML community, we are trying to move away from having to generate a large amount of data before we can see a significant impact of our ML methods. We believe that the improved signal to the HNU predictor (by penalizing the data-driven features of state quality and step progress upon hint usage) is significant enough to justify its use without collecting and learning from a new dataset. To assess whether the change in the signal because of incorporating hint usage is significant, we ran the Kolmogorov-Smirnov tests (KS test). We found that the distributions of these quality and progress features with and without penalties for hint usage are significantly different (\textit{p} $<$ .05). Further, these features contribute to about 95\% of the predictive power of the HN predictor \cite{maniktala2020jedm}. These results suggest that the significant change in the distributions of these features must have had a significant impact on the HNU predictor’s HelpNeed predictions. This analysis provides evidence to support that it is possible to develop an effective predictor of HelpNeed by only improving the signal, i.e. feedback about student behavior. That is, we don’t need to wait a semester for gathering data before we can have a significant impact on student learning.

\subsection{Limitations}
A limitation of this work is that we only assess our method in one ITS for one domain. As mentioned in our prior work \cite{maniktala2020jedm}, the requirements to apply our HelpNeed model in a new domain are to have state and state-transition representations (so we can define steps) and scoring for final solutions so the state-based and state-free classifiers can learn values for the HelpNeed predictor. Deriving state representations, and therefore, the HelpNeed methods, should be relatively straightforward in well-structured domains such as multi-step math, physics, or statistics problems. Further, our method for incorporating hint usage in predicting HelpNeed can be translated to such domains because they have problem-solving steps comprising two components: the next statement and its justification. However, studies are needed to confirm the effectiveness of our method in these domains.

Another limitation of this study is that the comparisons of help behaviors (possible avoidance, possible aptness, and possible abuse) rely on our HelpNeed classification and predictor, which are only shown to be correlated to posttest performance, but have not been proven to correspond directly to expert measures of help need. 

A final limitation of this study is the impact of the COVID-19 related lockdown. As mentioned earlier, the lockdown happened in the middle of the study which resulted in all the classes to go fully online. We had a lower assignment completion rate (88\%) than other semesters (above 95\%). While most COVID-19 lockdown related factors would have affected both the HNU conditions in a similar manner, there could have been some differences. For example, students in the HNU Adaptive condition may have become more dependent on hints to finish the assignment, causing a higher possible help abuse than expected, while students in the HNU Control condition may have been so focused on solving problems that they forgot to request help, causing a lower number of on-demand hints than the previous semester.

\section{Conclusion} 
This paper investigates a method to incorporate students’ hint usage  to more effectively determine when a student most needs assistance. Our HNU study suggests that incorporating student help usage behavior by simply including the number of hints used is not as useful as understanding and modifying the highly predictive features of the HelpNeed model (i.e., quality and progress features). We present a simple yet effective method to incorporate hint usage into our prior HelpNeed model \cite{maniktala2020jedm} to create a new HelpNeed and Use (HNU) model. Our empirical results suggest that incorporating hint usage into the model achieved our goal to significantly reduce incorrect HelpNeed predictions, meaning that we can provide more targeted help -- especially when it is needed. As a result, the HNU adaptive hint policy was successfully able to significantly reduce training unproductivity (HelpNeed steps) in the Adaptive condition when compared to the Control. In comparison with the HNU Control, the HNU Adaptive condition formed shorter, more efficient solutions faster and with higher accuracy, in the training section. This, in turn, enabled them to form shorter solutions with higher accuracy in the posttest (without hints). Furthermore, the adaptive hint policy helped save students in the HNU Adaptive condition about half an hour during training, in comparison with the Control, with the Adaptive group taking 64 min on average, and Control taking 90 min on average. These results suggest that our HNU model successfully addresses the assistance dilemma. Recall that, according to the assistance dilemma, withholding more information than needed can lead to frustration and wasted time, and giving more information than needed can lead to shallow learning and a lack of motivation to learn by oneself. The results of our experiment suggest that our method addresses the assistance dilemma because it prevents both of these negative impacts.

We further investigated the impact of the adaptive hint system through the lens of the assistance dilemma, looking at possible help avoidance, help appropriateness, and help abuse. We observe a significantly lower possible help avoidance, and significantly higher possible help appropriateness in the HNU Adaptive condition compared to \textit{not only} the HNU Control \textit{but also} the HN Adaptive condition (where HelpNeed predictions did not incorporate hint usage). However, we also observed significantly more possible help abuse in the HNU Adaptive condition than the HN Adaptive condition. While the COVID-19 related lockdown may have impacted these results, we list design changes that may help reduce this gaming behavior.

\section{Acknowledgements}
This material is based upon work supported by the National Science Foundation under Grant No. \#1726550 on Integrated Data-driven Technologies for Individualized Instruction in STEM Learning Environments.

\bibliographystyle{spmpsci}
\bibliography{ref}

\begin{thebibliography}{10}
\providecommand{\url}[1]{{#1}}
\providecommand{\urlprefix}{URL }
\expandafter\ifx\csname urlstyle\endcsname\relax
  \providecommand{\doi}[1]{DOI~\discretionary{}{}{}#1}\else
  \providecommand{\doi}{DOI~\discretionary{}{}{}\begingroup
  \urlstyle{rm}\Url}\fi

\bibitem{aleven2001helping}
Aleven, V.: Helping students to become better help seekers: Towards supporting
  metacognition in a cognitive tutor.
\newblock German-USA Early Career Research Exchange Program: Research on
  Learning Technologies and Technology-Supported Education, Tubingen, Germany
  (2001)

\bibitem{aleven2006toward}
Aleven, V., Mclaren, B., Roll, I., Koedinger, K.: Toward meta-cognitive
  tutoring: A model of help seeking with a cognitive tutor.
\newblock International Journal of Artificial Intelligence in Education
  \textbf{16}(2), 101--128 (2006)

\bibitem{arroyo2001analyzing}
Arroyo, I., Beck, J.E., Beal, C.R., Wing, R., Woolf, B.P.: Analyzing
  students’ response to help provision in an elementary mathematics
  intelligent tutoring system.
\newblock In: Papers of the AIED-2001 workshop on help provision and help
  seeking in interactive learning environments, pp. 34--46. Citeseer (2001)

\bibitem{azevedo2004does}
Azevedo, R., Cromley, J.G.: Does training on self-regulated learning facilitate
  students' learning with hypermedia?
\newblock Journal of educational psychology \textbf{96}(3), 523 (2004)

\bibitem{d2008more}
d~Baker, R.S., Corbett, A.T., Aleven, V.: More accurate student modeling
  through contextual estimation of slip and guess probabilities in bayesian
  knowledge tracing.
\newblock In: International conference on intelligent tutoring systems, pp.
  406--415. Springer (2008)

\bibitem{barnes2011using}
Barnes, T., Stamper, J., Croy, M.: "using markov decision processes for
  automatic hint generation".
\newblock Handbook of Educational Data Mining \textbf{467} (2011)

\bibitem{barnes2008pilot}
Barnes, T., Stamper, J.C., Lehmann, L., Croy, M.J.: A pilot study on logic
  proof tutoring using hints generated from historical student data.
\newblock In: EDM, pp. 197--201 (2008)

\bibitem{bartholome2006matters}
Bartholom{\'e}, T., Stahl, E., Pieschl, S., Bromme, R.: What matters in
  help-seeking? a study of help effectiveness and learner-related factors.
\newblock Computers in Human Behavior \textbf{22}(1), 113--129 (2006)

\bibitem{beck2008does}
Beck, J.E., Chang, K.m., Mostow, J., Corbett, A.: Does help help? introducing
  the bayesian evaluation and assessment methodology.
\newblock In: International Conference on Intelligent Tutoring Systems, pp.
  383--394. Springer (2008)

\bibitem{beck2013wheel}
Beck, J.E., Gong, Y.: Wheel-spinning: Students who fail to master a skill.
\newblock In: International conference on artificial intelligence in education,
  pp. 431--440. Springer (2013)

\bibitem{beck2004automatically}
Beck, J.E., Jia, P., Mostow, J.: Automatically assessing oral reading fluency
  in a computer tutor that listens.
\newblock Technology Instruction Cognition and Learning \textbf{2}, 61--82
  (2004)

\bibitem{bloom19842}
Bloom, B.S.: The 2 sigma problem: The search for methods of group instruction
  as effective as one-to-one tutoring.
\newblock Educational researcher \textbf{13}(6), 4--16 (1984)

\bibitem{bono2020stress}
Bono, G., Reil, K., Hescox, J.: Stress and wellbeing in urban college students
  in the us during the covid-19 pandemic: Can grit and gratitude help?
\newblock International Journal of Wellbeing \textbf{10}(3) (2020)

\bibitem{borek2009much}
Borek, A., McLaren, B.M., Karabinos, M., Yaron, D.: How much assistance is
  helpful to students in discovery learning?
\newblock In: European Conference on Technology Enhanced Learning, pp.
  391--404. Springer (2009)

\bibitem{botelho2019developing}
Botelho, A., Varatharaj, A., Patikorn, T., Doherty, D., Adjei, S., Beck, J.:
  Developing early detectors of student attrition and wheel spinning using deep
  learning.
\newblock IEEE Transactions on Learning Technologies  (2019)

\bibitem{brawner2011understanding}
Brawner, K.W., Holden, H.K., Goldberg, B.S., Sottilare, R.: Understanding the
  impact of intelligent tutoring agents on real-time training simulations.
\newblock Tech. rep., ARMY RESEARCH LAB ORLANDO FL HUMAN RESEARCH AND
  ENGINEERING DIRECTORATE (2011)

\bibitem{bunt2004scaffolding}
Bunt, A., Conati, C., Muldner, K.: Scaffolding self-explanation to improve
  learning in exploratory learning environments.
\newblock In: International Conference on Intelligent Tutoring Systems, pp.
  656--667. Springer (2004)

\bibitem{chaudhry2018modeling}
Chaudhry, R., Singh, H., Dogga, P., Saini, S.K.: Modeling hint-taking behavior
  and knowledge state of students with multi-task learning.
\newblock International Educational Data Mining Society  (2018)

\bibitem{conati2002using}
Conati, C., Gertner, A., Vanlehn, K.: Using bayesian networks to manage
  uncertainty in student modeling.
\newblock User modeling and user-adapted interaction \textbf{12}(4), 371--417
  (2002)

\bibitem{corbett1994knowledge}
Corbett, A.T., Anderson, J.R.: Knowledge tracing: Modeling the acquisition of
  procedural knowledge.
\newblock User modeling and user-adapted interaction \textbf{4}(4), 253--278
  (1994)

\bibitem{emerson2019predicting}
Emerson, A., Rodr{\'\i}guez, F.J., Mott, B., Smith, A., Min, W., Boyer, K.E.,
  Smith, C., Wiebe, E., Lester, J.: Predicting early and often: Predictive
  student modeling for block-based programming environments.
\newblock International Educational Data Mining Society  (2019)

\bibitem{feng2009addressing}
Feng, M., Heffernan, N., Koedinger, K.: Addressing the assessment challenge
  with an online system that tutors as it assesses.
\newblock User Modeling and User-Adapted Interaction \textbf{19}(3), 243--266
  (2009)

\bibitem{feng2006predicting}
Feng, M., Heffernan, N.T., Koedinger, K.R.: Predicting state test scores better
  with intelligent tutoring systems: developing metrics to measure assistance
  required.
\newblock In: International conference on intelligent tutoring systems, pp.
  31--40. Springer (2006)

\bibitem{fratamico2017applying}
Fratamico, L., Conati, C., Kardan, S., Roll, I.: Applying a framework for
  student modeling in exploratory learning environments: Comparing data
  representation granularity to handle environment complexity.
\newblock International Journal of Artificial Intelligence in Education
  \textbf{27}(2), 320--352 (2017)

\bibitem{hawkins2013extending}
Hawkins, W., Heffernan, N., Wang, Y., Baker, R.: Extending the assistance
  model: Analyzing the use of assistance over time.
\newblock In: Educational Data Mining 2013 (2013)

\bibitem{kai2018decision}
Kai, S., Almeda, M.V., Baker, R.S., Heffernan, C., Heffernan, N.: Decision tree
  modeling of wheel-spinning and productive persistence in skill builders.
\newblock JEDM| Journal of Educational Data Mining \textbf{10}(1), 36--71
  (2018)

\bibitem{kardan2015providing}
Kardan, S., Conati, C.: Providing adaptive support in an interactive simulation
  for learning: An experimental evaluation.
\newblock In: Proceedings of the 33rd Annual ACM Conference on Human Factors in
  Computing Systems, pp. 3671--3680. ACM (2015)

\bibitem{kinnebrew2014analyzing}
Kinnebrew, J.S., Segedy, J.R., Biswas, G.: Analyzing the temporal evolution of
  students’ behaviors in open-ended learning environments.
\newblock Metacognition and learning \textbf{9}(2), 187--215 (2014)

\bibitem{kock2010towards}
Kock, M., Paramythis, A.: Towards adaptive learning support on the basis of
  behavioural patterns in learning activity sequences.
\newblock In: 2010 International Conference on Intelligent Networking and
  Collaborative Systems, pp. 100--107. IEEE (2010)

\bibitem{koedinger2007exploring}
Koedinger, K.R., Aleven, V.: Exploring the assistance dilemma in experiments
  with cognitive tutors.
\newblock Educational Psychology Review \textbf{19}(3), 239--264 (2007)

\bibitem{ma2014intelligent}
Ma, W., Adesope, O.O., Nesbit, J.C., Liu, Q.: Intelligent tutoring systems and
  learning outcomes: A meta-analysis.
\newblock Journal of educational psychology \textbf{106}(4), 901 (2014)

\bibitem{maniktala2020extending}
Maniktala, M., Barnes, T., Chi, M.: Extending the hint factory: Towards
  modelling productivity for open-ended problem-solving.
\newblock In: In proceedings of the 13th International Conference on
  Educational Data Mining (DC paper) (2020)

\bibitem{maniktala2020leveraging}
Maniktala, M., Cody, C., Barnes, T., Chi, M.: Avoiding help avoidance: Using
  interface design changes to promote unsolicited hint usage in an intelligent
  tutor.
\newblock International Journal of Artificial Intelligence in Education
  \textbf{30}(4), 637--667 (2020)

\bibitem{maniktala2020jedm}
Maniktala, M., Cody, C., Isvik, A., Lytle, N., Chi, M., Barnes, T., et~al.:
  Extending the hint factory for the assistance dilemma: A novel, data-driven
  helpneed predictor for proactive problem-solving help.
\newblock Journal of Educational Data Mining \textbf{12}(4), 24--65 (2020)

\bibitem{mayer1992thinking}
Mayer, R.E.: Thinking, problem solving, cognition.
\newblock WH Freeman/Times Books/Henry Holt \& Co (1992)

\bibitem{mclaren2008and}
McLaren, B.M., Lim, S.J., Koedinger, K.R.: When and how often should worked
  examples be given to students? new results and a summary of the current state
  of research.
\newblock In: Proceedings of the 30th annual conference of the cognitive
  science society, pp. 2176--2181 (2008)

\bibitem{mclaren2014web}
McLaren, B.M., Timms, M., Weihnacht, D., Brenner, D., Luttgen, K., Grillo-Hill,
  A., Brown, D.H.: A web-based system to support inquiry learning.
\newblock In: Proceedings of the 6th International Conference on Computer
  Supported Education-Volume 1, pp. 43--52. SCITEPRESS-Science and Technology
  Publications, Lda (2014)

\bibitem{mostafavi2015data}
Mostafavi, B., Zhou, G., Lynch, C., Chi, M., Barnes, T.: Data-driven worked
  examples improve retention and completion in a logic tutor.
\newblock In: International Conference on Artificial Intelligence in Education,
  pp. 726--729. Springer (2015)

\bibitem{murray2005effects}
Murray, R.C., VanLehn, K.: Effects of dissuading unnecessary help requests
  while providing proactive help.
\newblock In: AIED, pp. 887--889. Citeseer (2005)

\bibitem{murray2006comparison}
Murray, R.C., VanLehn, K.: A comparison of decision-theoretic, fixed-policy and
  random tutorial action selection.
\newblock In: International Conference on Intelligent Tutoring Systems, pp.
  114--123. Springer (2006)

\bibitem{murray2004looking}
Murray, R.C., Vanlehn, K., Mostow, J.: Looking ahead to select tutorial
  actions: A decision-theoretic approach.
\newblock International Journal of Artificial Intelligence in Education
  \textbf{14}(3, 4), 235--278 (2004)

\bibitem{murray2003overview}
Murray, T.: An overview of intelligent tutoring system authoring tools: Updated
  analysis of the state of the art.
\newblock In: Authoring tools for advanced technology learning environments,
  pp. 491--544. Springer (2003)

\bibitem{pena2011improving}
Pe{\~n}a, A., Kayashima, M., Mizoguchi, R., Dominguez, R.: Improving
  students’ meta-cognitive skills within intelligent educational systems: A
  review.
\newblock In: International Conference on Foundations of Augmented Cognition,
  pp. 442--451. Springer (2011)

\bibitem{polya2004solve}
Polya, G.: How to solve it: A new aspect of mathematical method, vol.~85.
\newblock Princeton university press (2004)

\bibitem{razzaq2010hints}
Razzaq, L., Heffernan, N.T.: Hints: is it better to give or wait to be asked?
\newblock In: International Conference on Intelligent Tutoring Systems, pp.
  349--358. Springer (2010)

\bibitem{smith2012toward}
Smith, M.U.: Toward a unified theory of problem solving: Views from the content
  domains.
\newblock Routledge (2012)

\bibitem{son2020effects}
Son, C., Hegde, S., Smith, A., Wang, X., Sasangohar, F.: Effects of covid-19 on
  college students’ mental health in the united states: Interview survey
  study.
\newblock Journal of medical internet research \textbf{22}(9), e21279 (2020)

\bibitem{stone1993missing}
Stone, C.A.: What is missing in the metaphor of scaffolding.
\newblock Contexts for learning: Sociocultural dynamics in children’s
  development pp. 169--183 (1993)

\bibitem{tchetagni2002hierarchical}
Tch{\'e}tagni, J.M., Nkambou, R.: Hierarchical representation and evaluation of
  the student in an intelligent tutoring system.
\newblock In: International Conference on Intelligent Tutoring Systems, pp.
  708--717. Springer (2002)

\bibitem{ueno2017irt}
Ueno, M., Miyazawa, Y.: Irt-based adaptive hints to scaffold learning in
  programming.
\newblock IEEE Transactions on Learning Technologies \textbf{11}(4), 415--428
  (2017)

\bibitem{vanlehn2011relative}
VanLehn, K.: The relative effectiveness of human tutoring, intelligent tutoring
  systems, and other tutoring systems.
\newblock Educational Psychologist \textbf{46}(4), 197--221 (2011)

\bibitem{vygotsky1980mind}
Vygotsky, L.S.: Mind in society: The development of higher psychological
  processes.
\newblock Harvard university press (1980)

\bibitem{wang2011assistance}
Wang, Y., Heffernan, N.T.: The" assistance" model: Leveraging how many hints
  and attempts a student needs.
\newblock In: FLAIRS Conference (2011)

\bibitem{wang2010representing}
Wang, Y., Heffernan, N.T., Beck, J.E.: Representing student performance with
  partial credit.
\newblock In: EDM, pp. 335--336. Citeseer (2010)

\bibitem{wood1976role}
Wood, D., Bruner, J.S., Ross, G.: The role of tutoring in problem solving.
\newblock Journal of child psychology and psychiatry \textbf{17}(2), 89--100
  (1976)

\end{thebibliography}

\end{document}